# 3D Topology Optimization Using Convolutional Neural Networks

Saurabh Banga, Harsh Gehani, Sanket Bhilare, Sagar Jitendra Patel, Levent Burak Kara[1]

*Mechanical Engineering*
*Carnegie Mellon University*
*Pittsburgh, USA*

*Abstract-* **Topology optimization is computationally demanding that requires the assembly and solution to a finite element problem for each material distribution hypothesis. As a complementary alternative to the traditional physics-based topology optimization, we explore a data-driven approach that can quickly generate accurate solutions. To this end, we propose a deep learning approach based on a 3D encoder-decoder Convolutional Neural Network architecture for accelerating 3D topology optimization and to determine the optimal computational strategy for its deployment. Analysis of iteration-wise progress of the Solid Isotropic Material with Penalization process is used as a guideline to study how the earlier steps of the conventional topology optimization can be used as input for our approach to predict the final optimized output structure directly from this input. We conduct a comparative study between multiple strategies for training the neural network and assess the effect of using various input combinations for the CNN to finalize the strategy with the highest accuracy in predictions for practical deployment. For the best performing network, we achieved about 40% reduction in overall computation time while also attaining structural accuracies in the order of 96%.**

*Keywords— Deep learning, data-driven 3D topology optimization, Convolutional Neural Networks*

1. **INTRODUCTION**

Topology optimization generates structures by optimizing the material distribution inside a design domain subject to specified loads and constraints. With the emergence of additive manufacturing capable of producing complex structures, topology optimization has become increasingly attractive, taking a central role in today's many structural design problems. Some of these have been discussed by Paris *et al*. [1], Kingman *et al*. [2] and Wang *et al*. [3] among many others.

At the heart of topology optimization is the computationally demanding finite element analysis that evaluates the prescribed objective function(s) to iteratively guide the material distribution process (Bendsøe [4], Zhou and Rozvany [5], Liu and Tovar [6], Andreassen *et al*. [7] and Hunter [8]). Muller *et al*. [9] describe the high computational requirements for large scale topology optimization. A comparative study of commercial topology optimization tools and the large computational time required for the same has also been presented by Wook-han *et al*. [10]. While significant advances have been made in efficiently computing the optimal solutions ([6] and Suresh K [11]), the overall computational complexity, necessity of high performance computation resources and significant time requirements still present a major challenge in a large-scale deployment of topology optimization.

In this work, we aim to complement traditional topology optimization with a data-driven approach as a way to accelerate the search for the optimized structure. Our approach rests on the theoretical ideal that with a sufficiently broad set of data that spans variations in the loads, boundary conditions, materials, objectives and design domains, a regressor with enough degrees of freedom can be trained to establish a mapping from the input to the optimized structures. Clearly, the space of variations that must be covered in this way is infinitely large. Nonetheless, in this work, we aim to establish the foundation for such an approach with simplifications that enable various important data sources to be parametrically studied. To this end, we use a 3D convolutional neural network (CNN) that can take as input intermediate solutions to the material distribution and predict the final structure. The network is trained on a fixed discretized domain and a single material type, while varying the location, number, and magnitude of the external point forces, and the nature of the displacement boundary conditions. With these variations, we explore whether the earlier stages of the conventional topology optimization could be leveraged to estimate the resulting optimized structures under a number of test scenarios.

Once trained, the prediction of the final structure using the CNN is virtually instantaneous. As such, this utility can be seen as an approach that exchanges the cost of a large corpus of off-line training data and potential inaccuracy against run-time performance. Nonetheless, the proposed method cannot and does not substitute conventional topology optimization. Instead, it seeks to establish a framework that can alleviate the need for many finite element calculations to generate solutions that are either similar to those generated by the conventional optimizer, or at least serve as good starting points for downstream conventional iterations.

---

[1] Corresponding author, lkara@cmu.edu

Declarations of interest: None

NOTE: THIS MANUSCRIPT INVOLVED EQUAL WORK CONTRIBUTION BY THE AUTHORS: 1,2,3,4 FOR JOINT FIRST-AUTHORSHIP

We explicitly limit our scope to density-based compliance minimization problems where the design domain is parametrized with pseudo-densities that serve as a discrete representation of a continuous field of Young's moduli. In such formulations such as the Solid Isotropic Material with Penalization (SIMP) method, the optimizer seeks to minimize the compliance of the structure subject to a specified volume fraction (enforced as an explicit constraint or penalty in the objective function). Based on the observation that the structure evolves as the result of the solver following the gradients dictated by the objective function(s) and the associated material and equilibrium constraints, we explore whether the knowledge gained during the earlier iterations of such methods can be utilized to directly estimate the final optimized structures.

2. **RELATED WORK**

Topology optimization is evolving rapidly. A thorough review of current and future trends in the field has been presented by Liu *et al.* [12], Mirzendehdel and Suresh [13] where they discuss the use of topology optimization targeting additive manufacturing. The work discusses developments and challenges of using topology optimization across variety of application involving support structure designs, lattice structure generation, structural design with material or manufacturing uncertainties among others. Biryikli and To [14] have proposed a non-sensitivity method termed as Proportional Topology Optimization to solve the stress constrained and minimum compliance problems which is the test case for our research as well. An alternative approach to tackle the problem based on level-set defined via the topological derivative has also been presented by Suresh and Takalloozedh [15]. Da *et al.* [16] and Xia *et al.* [17] have shown how the Bi-directional Evolutionary Structural Optimization (BESO) method can be further improved for the generation of better structures as part of an evolutionary topology optimization strategy. The objective is to increase the robustness and the effectiveness of the process while generating structures with smooth boundary representation and less dependency on the finite element mesh resolution. Similarly, Lambe and Czekanski [18] have presented how process improvements like using continuous density field and adaptive mesh refinement can help resolve many of the numerical issues associated with topology optimization like checkerboard patterns and material islands. Furthermore, Liu and To [19] and Mirzendehdel and Suresh [20] have also discussed different algorithms and methodologies for topology optimization in additive manufacturing areas like support structure optimization and hybrid manufacturing for designs with geometric complexity as well as high dimensional accuracy. Deng and Suresh [21][22] have provided guidelines to stress constrained thermo-elastic topology optimization specifically for problems related to buckling and variable temperature fields. They also discuss how topologies for pure-elastic problem vary from those for thermo-elastic problems.

Although new theory and techniques are continuously being developed, the problem itself remains complex and computationally demanding. While acknowledging the groundbreaking research which has taken place in this field over the past many years, this paper has not targeted and attempted to modify the specifics of the widely-accepted state-of-the art methodologies. This work adds on to these methods in an out-of-the-box approach with an underlying goal of making the process more efficient. We propose a deep learning framework that is largely dependent on the data and outputs received from the conventional processes. Our work thus is inherently dependent on the robustness of these solvers and methods.

Attempts have also been made to tackle the problem of topology optimization using a data-driven approach. Works from Bobby *et al.* [23] as well as Ulu *et al.* [24] have explored the application of such techniques. They also discuss the challenges like inaccurate or sub-optimal structural predictions associated with the data-driven techniques and propose novel solutions. Ulu *et al.* [25] have shown how a critical instant analysis method can be used to solve the optimization problem of lightweight structure design under force location uncertainty.

A more resource-centric approach is to use computational capabilities of a Graphics Processing Unit (GPU) for large scale topology optimization, as proposed in multiple papers by Martínez-Frutos *et al.* [26], Martínez-Frutos and Herrero-Pérez [27] and Challis *et al.* [28]. These works have elucidated the use of parallel GPU implementation to achieve extremely high resolution and accuracy in structures while accelerating the overall process. On similar lines, the use of machine learning can also prove to be very effective in overcoming the challenges of high computation times and requirement of large resources for topology optimization.

The use of machine learning techniques for topology optimization is relatively new but it has gained traction in the past few years. Yu *et al.* [29] use generative modeling techniques for topology optimization. They use variational auto encoders (VAE) and Generative Adversarial Networks (GANs) to predict the optimized structure after topology optimization. The inputs in the form of loading and displacement boundary conditions are provided to their generative networks.

The paper by Ulu *et al.* [24] proposes the use of Principal Component Analysis (PCA) and a fully connected neural network to learn the mapping between loading configurations and optimal topologies. In our research, although we also explore the incorporation of loading and boundary configurations as one of the possible inputs, we use a CNN instead, to converge to an optimal 3D topology from an intermediate 3D topology.

Sosnovik and Oseledets [30] use convolutional encoder-decoder architecture to solve the topology optimization problem by posing it as an image segmentation task. Our approach is similar to theirs in that we have also use deep learning to transform an intermediate topology to the optimal topology. However, their work is only limited to 2D structures and only takes into consideration the density distribution and the gradient of the density distribution as the inputs to train the model.

In this paper, we take it one step further and experiment with incorporation of loads and boundary conditions as viable inputs as well. We also present an extensive study on mapping of the topology optimization process which can be used to select a potential point of transfer between the conventional method and the deep learning framework. This makes the results of this paper more practical from implementation point-of-view. Additionally, a more stringent accuracy metric (root mean squared accuracy) has also been introduced to evaluate the performance of the convolutional neural network for topology optimization. This helps in understanding the proximity of the actual density distribution values between the predicted and the final structure, as opposed to only evaluating binary structures obtained after thresholding. Although the work by [30] demonstrates a good proof-of-concept, it does not fully touch upon its practicality and discuss the critical choices which need to be made during deployment of the proposed methodology. Our work sufficiently extends the research and provides guidelines for the most optimum computational strategy by presenting an extensive comparative analysis. Moreover, all the aforementioned papers deal with 2D topology optimization, while our research explores topology optimization of 3D structures using deep learning approaches.

3. **THEORY**

We aim to learn the solutions to the problem of minimum compliance topology optimization and limit our discussions to the solutions obtained using the Solid Isotropic Material with Penalization (SIMP) method. In SIMP, the objective is to find the material density distribution of physical densities **x** such that the strain energy is minimized under the prescribed support and loading conditions.

The decision (design) variables comprise of the actual density distribution vector $\mathbf{x} = [x_1, x_2, \ldots, x_n]^T$ where n is the number of elements. The physical density distribution **x** is mapped to modified densities $\tilde{\mathbf{x}}$ using a neighborhood density filter function. The density filter function is added to smoothen the discretized original density distribution. It is expressed as:

$$\text{Density Filter Function: } \tilde{x}_i = \frac{\sum_{j \in N_i} h_{ij} v_j x_j}{\sum_{j \in N_i} h_{ij} v_j}$$

Here Neighborhood $N_i = \{j: dist(i,j) \leq r_{MIN}\}$, where $dist(i,j)$ is the distance between the center of element $i$ and the center of element $j$. $r_{MIN}$ is the size of the neighborhood or filter size. Weight factor $h_{ij} = r_{MIN} - dist(i,j)$, where $j \in N_i$

The brief mathematical formulation is defined as follows:

| | | |
|---|---|---|
| Minimize the compliance of a mechanical structure | : - | $\underset{\{\mathbf{x} \in \mathbb{R}^n\}}{minimize} \ f(\tilde{\mathbf{x}}) = \mathbf{u}(\tilde{\mathbf{x}})^T \mathbf{K}(\tilde{\mathbf{x}}) \mathbf{u}(\tilde{\mathbf{x}})$ |
| **Subject to** | | |
| Internal reactive forces balance the external forces | : - | $h(\tilde{\mathbf{x}}) = \mathbf{K}(\tilde{\mathbf{x}})\mathbf{u}(\tilde{\mathbf{x}}) - \mathbf{f} = \mathbf{0}$ |
| Total volume should be less than or equal to the desired volume | : - | $g(\tilde{\mathbf{x}}) = \tilde{\mathbf{x}}^T \mathbf{v} - v_0 V \leq 0$ |
| Bounds | : - | $\tilde{\mathbf{x}}_{LB} \leq \tilde{\mathbf{x}} \leq \tilde{\mathbf{x}}_{UB}$ <br> ($\tilde{\mathbf{x}}_{LB}, \tilde{\mathbf{x}}_{UB} : \mathbf{0}, \mathbf{1}$ respectively) |

Where,
Design Variables : $\mathbf{x} = [x_1, x_2, ..., x_n]^T$
$\tilde{\mathbf{x}} = \phi(\mathbf{x})$ : Vector of filtered element densities
$\mathbf{v} = [v_1, v_2, ..., v_n]^T$ : Vector of element volume
$\mathbf{K}_i^0$ : Element stiffness matrix (constant)
$\mathbf{K}_i(\tilde{\mathbf{x}})$ : Global element stiffness matrix
$((e_0 + (e_0 - e_{MIN})(\widetilde{\mathbf{x}_i})^p)\mathbf{K}_i^0$

$e_0$ : Material Young's Modulus
$e_{MIN}$ : Arbitrary finite positive value
$p$ : Penalization factor
$\mathbf{K}(\tilde{\mathbf{x}})$ : Assembled Global Stiffness Matrix
$\mathbf{u}(\tilde{\mathbf{x}})$ : Vector of Nodal Displacements
$\mathbf{f}$ : Force Vector
$v_0$ : Prescribed volume fraction
$V$ : Volume of the original design

The iterative progress of the SIMP method for a sample case is depicted in Figure 1 (only voxels with material density values above 0.5 are shown to illustrate the structural outline). The material density distribution is observed to attain high spatial resolution early in the process with slow progress subsequently.

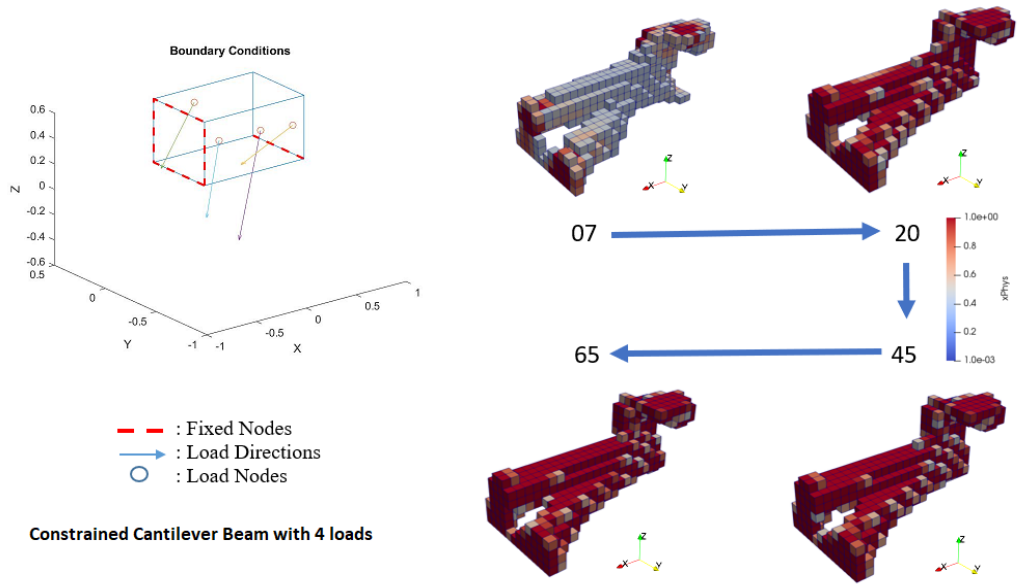

Figure 1: Iterative Progress of SIMP-based 3D Topology Optimization (Domain: 24 x 12 x 12)

4. **METHODS**

We propose a data-driven approach to topology optimization to complement a traditional approach and aim to reduce the time required to arrive at the optimized 3D structure through the use of a 3D convolutional network architecture as a regressor. The initial iterations of topology optimization are performed using an FEA-based software 'TopOpt' [31] which is based on the SIMP method. In our methodology, we stop the topology optimization solver at an intermediate stage and then feed the current structure from the optimizer to our network which predicts the final optimized structure. Since our network is trained using the results of TopOpt [31] as the ground truth, our approach does not provide a pathway to finding pareto-optimal solutions for cases where the volume fraction is utilized as a penalty in the objective, or to identify solutions of equal quality but geometrically distant solutions.

Furthermore, to construct and test our approach, we limit our domain to 3D Messerschmitt-Bölkow-Blohm (MBB) Beam subject to four different subtypes of this general beam model and the parametric variations on the forces and boundary conditions for each subtype.

## 4.1 Data Generation

We generated synthetic data using the open source topology optimization tool 'TopOpt' [31], which works for structural problems pertaining to SIMP methodology. The specifications of the topology optimization problem defined for this work are summarized below:

Domain type             : Rectangular Beam
Domain Size             : 24 x 12 x 12
Beam Dimensions         : 2 m x 1m x 1m
Minimum Element size    : 0.083 m
No. of Elements         : 3456

To generate spatially variant data, we devised a sampling strategy to define the loading and boundary conditions for the topology optimization problem. Parameters which are sampled are depicted in Figure 2.

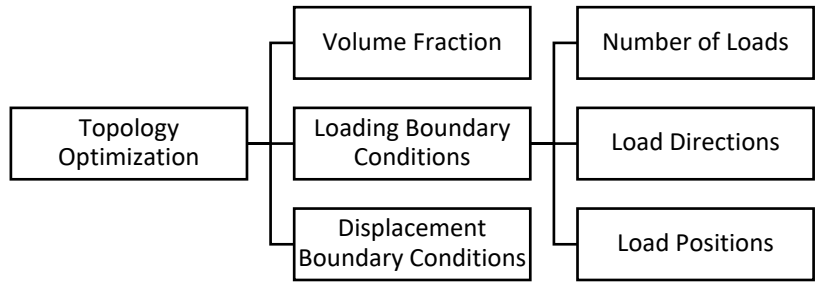

Figure 2: Parameters sampled for data generation

Sampling works as follows:
- Volume fraction ($V_0$)
  $V_0$: Sampled from Normal Distribution $\mathcal{N}(\mu=0.28; \sigma=0.07)$
  The above parameters are chosen to ensure that volume fraction values primarily vary between 0.07 to 0.5. Volume fraction is limited to lower values (below 0.5) to ensure high variation in material distribution within the design domain. Convergence for topology optimization problems with low volume fraction is known to be computationally complex also resulting in high simulation times.

- Number of load Nodes ($N_L$)
  $N_L$: Sampled from Poisson Distribution $P(\lambda=4)$
  The above parameters are chosen to ensure that number of loads varied from 1 to 10. Here 4 (mean) is the most probable value while the probability of sampling decreases as you move away from it. As we apply distributed loads around every load node, we want to avoid overcrowding of load nodes in the structure while also ensuring that sufficient number of loads are indeed acting.

- Load direction vectors ($\mathbf{V}_L$)
  $\mathbf{V}_L$: Uniform Random Sampling (0:1) across x, y, z directions
  Each load direction is normalized to derive the unit vector direction. The strategy of sampling across all three directions is because having only unidirectional or axial loads poses the possibility of creating redundancy in the structures. Also, by choosing unit vectors we ensure that the actual total magnitude of loads acting on the structure is governed implicitly by the number of load nodes ($N_L$) itself.

- Load node positions ($P_L$)
  Positions of nodes for application of the loads are randomly sampled from the 6 boundary faces of the rectangular beam domain. This has been done to mimic practical loading conditions.

  One of the boundary planes is randomly selected and the remaining two co-ordinates are sampled using the following strategy for each node:
  $P_L$ (x): Uniform random sampling [0: 1]
  $P_L$ (y): Uniform random sampling [0: 0.5]
  $P_L$ (z): Uniform random sampling [0: 0.5]

The applied load is distributed over a hemi-spherical domain (radius = 1 x element size) around each load node (on the boundary of the beam). This avoids singularities associated with point-loads.

Standard load magnitude ($F_0$) of 1 or -1 (uniform random sampling) is chosen. Final load vector (**F**, a Matrix of size 3 x $N_L$) can be derived from **F** = $F_0$ x **$V_L$**

- Displacement Boundary Constraint case ($B_C$)
  $B_C$: Discrete random sampling from [1,2,3,4] to define constraint case

  Four typical real-world scenarios, as depicted in Figure 3 below, were used to sample the displacement boundary constraint cases:
  1. Cantilever Beam
  2. Simply Supported Beam
  3. Modified Simple Supported Beam
  4. Constrained Cantilever Beam

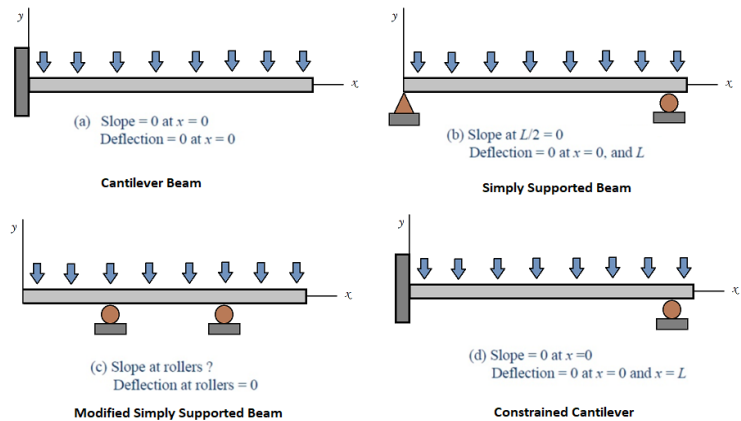

Figure 3: Constraint Boundary Condition Sample Cases

Based on the sampling of the boundary constraints, we generated a set of 6000 problem statements. Each of these data samples took about 70-100 iterations in the conventional topology optimization process to converge to the optimized output. Hence on an average about 6000 x 85 number of 3D material density distribution arrays of size 24 x 12 x 12 – each representing various structural configurations within the design domain- were generated for this work. Figure 4 shows an example of sampled boundary condition and the corresponding topological structures.

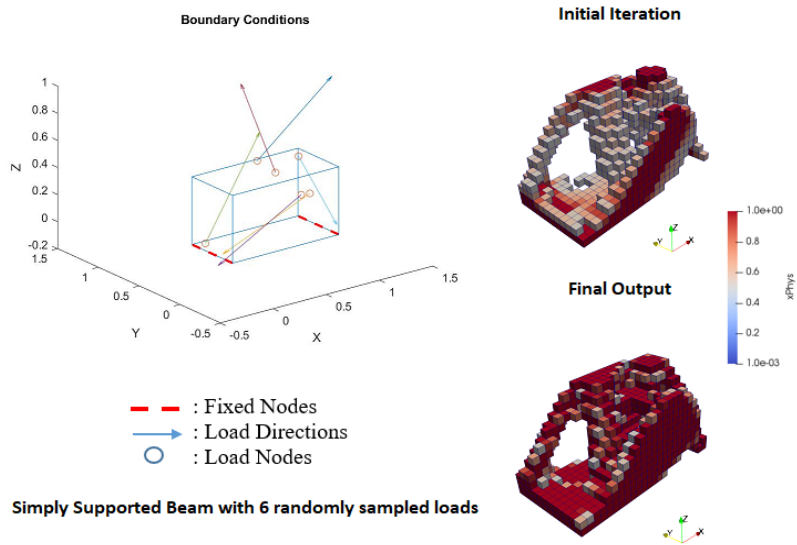

Figure 4: Sample cases for constraint boundary conditions chosen for data generation. Dark colors represent high density values closer to one, while the light colors represent the density values closer to zero

## 4.2 Network Architecture

Our network consists of only convolution layers without any dense layers as can be seen in Figure 5. 3D convolution with Rectified Linear Unit (ReLU) non-linear activation function between layers and MaxPool are utilized to down-sample the input in the encoder part of the network. The hyperbolic tangent (tanh) function is set as the output layer activation function since it resulted in the highest accuracy of all the activation functions with which we experimented. The network performs transpose convolution and complement it with convolution in the decoder part to get the final size of the output same as the input.

The loss function comprises of a confidence loss and a penalization term for enforcing volumetric constraint. Computation of the confidence loss with respect to the ground truth obtained from the TopOpt [31] solver is in the form of Binary Cross Entropy (BCE). The volumetric equality between the network output and ground truth is expressed as mean squared error. The mathematical formulation for the loss is expressed below:

$$\text{Loss} = \frac{-1}{n} \left[ \sum_{i=1}^{n} X_{true}^{i} \log(X_{pred}^{i}) + (1 - X_{true}^{i}) \log(1 - X_{pred}^{i}) \right] + \beta \left[ \frac{1}{n} \sum_{i=1}^{n} (X_{pred}^{i} - X_{true}^{i})^2 \right]$$

Stochastic gradient descent with momentum is used as the optimizer.

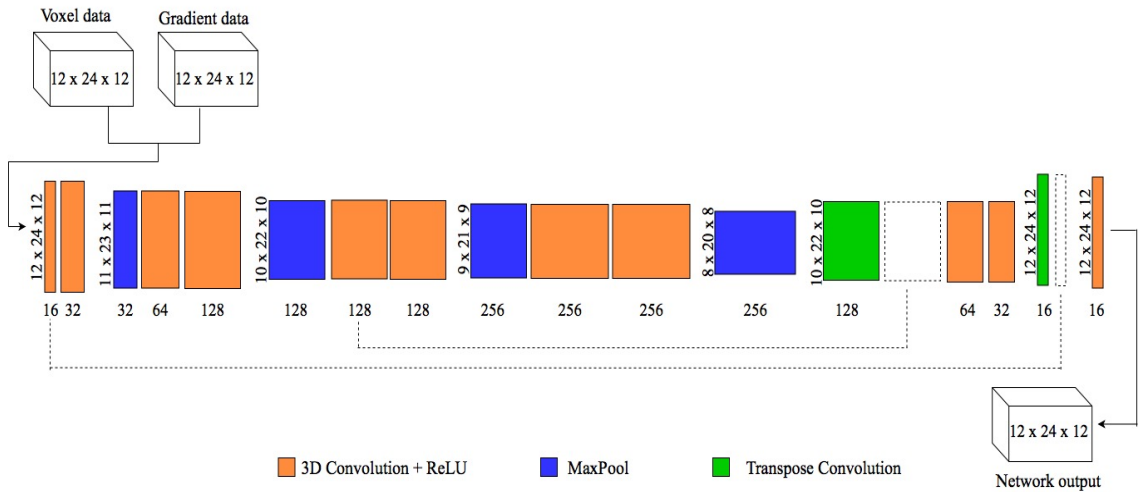

Figure 5: The Network Architecture depicts the input as concatenation of 3D voxel data and gradient data. For each layer, the dimensions of 3D data are depicted following notation of Height x Length x Width. Numbers below the boxes denote the number of channels.

## 4.3 Training the Network

Out of the 6000 data samples generated using the sampling strategy, we used 4500 data samples for training the network, 500 data samples formed cross-validation set for hyperparameter tuning decisions and the remaining 1000 samples were reserved as test data for assessing the generalized performance of the network. The hyperparameters chosen for the final network were obtained on the basis of their performance on the cross-validation set.

The network can learn from the density distribution and how the density varies. The network can also learn from the force and boundary condition for the problem. In addition to the intermediate density distribution data, we also provided the gradient information as it captures the details of how the structure changes between different iterations and helps the network understand this trend to converge to an appropriate output structure. Forces and boundary conditions are taken as inputs as they directly correlate with the overall structural skeleton. We wanted to see if the network learns something meaningful from them and assess whether it can predict an optimized structure using generative capabilities.

We trained multiple networks using different combinations of input parameters (channels) as described in Table 1, in an attempt to find out the best performing network details of which are discussed in the results section.

This led to 3 types of inputs for the convolutional neural network:
1. 3D density distribution of voxels at iteration *m* (*m* < *T*)
2. Gradient of voxel densities between iterations *m* and *n* (*n* < *m*)
3. Forces and Boundary Conditions along x, y and z directions

Here, *T* is the total number of iterations of a particular data sample while '*n*' and '*m*' are intermediate iterations. All the inputs are 3D matrices of size 24x12x12.

Input **1** is a 3D matrix where the values in the matrix are the density values at the corresponding voxels from the $m^{th}$ iteration. Input **2** is a 3D matrix where the values in the matrix are the differences in the density values between the $m^{th}$ and $n^{th}$ iterations at each voxel. In Input **3**, for the forces, we have three 3D matrices that specify the magnitude of the resultant force component acting at each voxel in the x, y and z directions respectively. Similarly, for the boundary conditions input, we have three 3D matrices for displacement constraints in x, y and z directions. All the voxels with constraints are assigned a value of 1 and those which are not constrained have a value of 0. Moreover, we incorporate the design domain boundary by assigning a value of -1 for voxels along the beam surface that are not constrained. So, a total of eight 3D matrices or channels which can be concatenated along a 4th dimension making the maximum size of the input to the 3D CNN as 24x12x12x8. Table 1 below gives mathematical description for each channel.

Table 1: Mathematical description for each possible channel to the network

| ($MN_{ijk}$ is the matrix at voxel with indices i, j and k. Here N (Channel index) = [1:8], while i = [0:24], j = [1:12] & k= [1:12]) | | |
|---|---|---|
| Channel 1 (Density) | $M1_{ijk} = (D_{xyz})_m$ | $D_{xyz}$ is the density value of voxel centered at x, y and z |
| Channel 2 (Density Gradient) | $M2_{ijk} = (D_{xyz})_m - (D_{xyz})_n$ | $M2_{ijk}$ is the element-wise difference between densities |
| Channel 3 (Force along X direction) | $M3_{ijk} = (FX_{xyz})$ | $FX_{xyz}$ is the force value in X direction at voxel centered at x, y and z |
| Channel 4 (Force along y direction) | $M4_{ijk} = (FY_{xyz})$ | $FY_{xyz}$ is the force value in Y direction at voxel centered at x, y and z. |
| Channel 5 (Force along z direction) | $M5_{ijk} = (FZ_{xyz})$ | $FZ_{xyz}$ is the force value in Z direction at voxel centered at x, y and z. |
| Channel 6 (Constraints in x direction) | $M6_{ijk} = (CX_{xyz})$ | $CX_{xyz}$, $CY_{xyz}$, $CZ_{xyz}$ are the constraints in x, y and z directions respectively. |
| Channel 7 (Constraints in y directions) | $M7_{ijk} = (CY_{xyz})$ | *C = 1* for constraint voxels<br>*C = 0* for unconstraint voxels |
| Channel 8 (Constraints in z direction) | $M8_{ijk} = (CZ_{xyz})$ | *C = -1* for unconstraint voxels on the boundary |

To ascertain how different intermediate points of transfer affect the performance of the network, we sampled the iteration number '*m*' using the (1) Poisson Distribution, and (2) Uniform Distribution. We created 4 different datasets based on different techniques for sampling '*m*' to assess how the network learns from input density distribution from different stages of the SIMP solver iterations.
- Data set 1: '*m*' is sampled using Uniform Distribution from 1 to *T*
- Data set 2: '*m*' is sampled using Poisson's Distribution with value of mean parameter (λ = 5)
- Data set 3: '*m*' is sampled using Poisson's Distribution with value of mean parameter (λ = 10)
- Data set 4: '*m*' is sampled using Poisson's Distribution with value of mean parameter (λ = 30)

These diverse datasets were finalized to make the network assessment meaningful. If we use data set 1 (uniform distribution) for training the network, the input density iteration *m* could be anything between 0 and the last iteration. For the 2nd data set (Poisson's distribution, λ = 5), the input density iteration will be from the earlier stages of iterations. For the 3rd data set (Poisson's distribution, λ = 10), the input density distribution will be from the midway stages of iterations. Lastly, for the 4th data set (Poisson's distribution, λ = 30), the input density distribution will be from the final stages of the iterations and hence will be the closest to convergence. Once '*m*' is sampled, we sample '*n*' using uniform distribution between 0 and *m*. We trained the network on all of these different data sets and have evaluated its performance. This is discussed comprehensively in the results section.

### 4.4 Process Mapping

To define a suitable cutoff point between the FEA solver and the 3D CNN for the proposed methodology, it is important to quantify the trends in the spatial variation of structures from iteration to iteration. To understand this progress of the topology optimization process, we map the change in material density against the number of iterations. Primarily two metrics are found to be relevant for mapping the progress: (1) Binary accuracy (2) Spatial gradient of filtered densities. To capture the trends, the study was conducted on 1000 random test samples from the total data generated. For each sample we have material density distribution data for each iteration, as each of the sample undergoes optimization within 70-100 number of iterations prior to convergence which is dependent on problem complexity.

#### 4.4.1 Binary Accuracy:

Binary accuracy is computed by comparing the material density distribution for intermediate structures after each iteration against the final density distribution in the optimized structure (ground truth). Thresholding of the density distribution is done to get the binary data. Detailed mathematical formulation for computing this metric is presented in Results and Discussion (section 5). The normalized trend for 1000 samples depicted in Figure 6, shows the variance of binary accuracy with respect to percentage progress in terms of iterations.

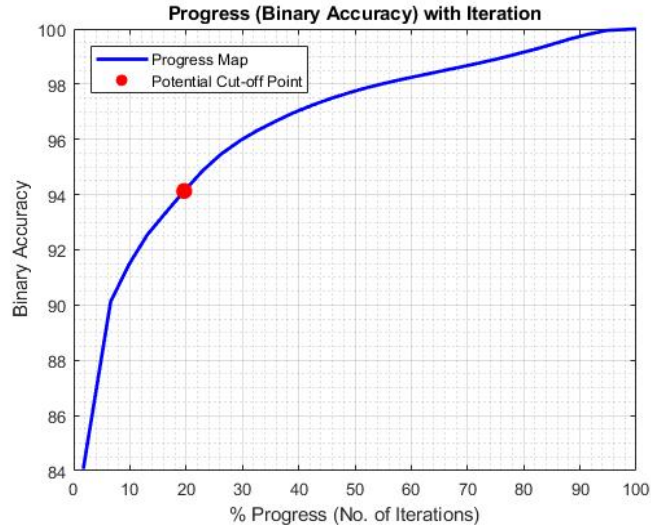

Figure 6: Binary Accuracy Vs Progress (1000 samples)

The topology optimization algorithm is observed to reach high binary accuracy values of greater than 90% within the first fifth of the process progress. As the algorithm progresses, the effective increase in accuracy is limited while time taken for the same is significantly large. The Convolutional Neural Network approach presented in this paper is deployed after such an initial resolution of the structure to drive the binary accuracy values higher without expending much computational time.

#### 4.4.2 Gradient Norm of neighborhood filtered density distribution:

We observed that the trend in the norm of the direct gradient of the material density values was not very useful (refer Appendix B). We found it to be continuously decreasing- in line with the standard gradient-based numerical algorithms typically used for topology optimization. Hence, for understanding the spatial change in density values with respect to iterations, we devised a modified gradient metric. For this metric, a neighborhood density filter was applied at each voxel. This density filter takes a weighted average of the density distribution in the neighborhood of each voxel. We then subtract this mean value from the actual density values at each voxel to compare how it varies with respect to its neighborhood. Hence for each iteration, we have a spatial map of the structure depicting the local variations in the density values inside the design domain. The metric is summarized below:

$$\text{Density Filter Function: } \widetilde{x}_i = \frac{\sum_{j \epsilon N_i} h_{ij} v_j x_j}{\sum_{j \epsilon N_i} h_{ij} v_j}$$

With Neighborhood $N_i = \{j: dist(i,j) \leq r_{MIN}\}$, where $dist(i,j)$ is the center distance between elements $i$ and $j$. $r_{MIN}$ is the size of the neighborhood. Weight factor, $h_{ij} = r_{MIN} - dist(i,j)$, where $j \in N_i$

$$\text{Spatial Map: } x_i^{'} = x_i - \widetilde{x}_i$$

The Frobenius norm of the iteration-wise gradient of the individual 3D arrays depicting local spatial variance (spatial maps) is mapped against the percentage progress in Figure 7.

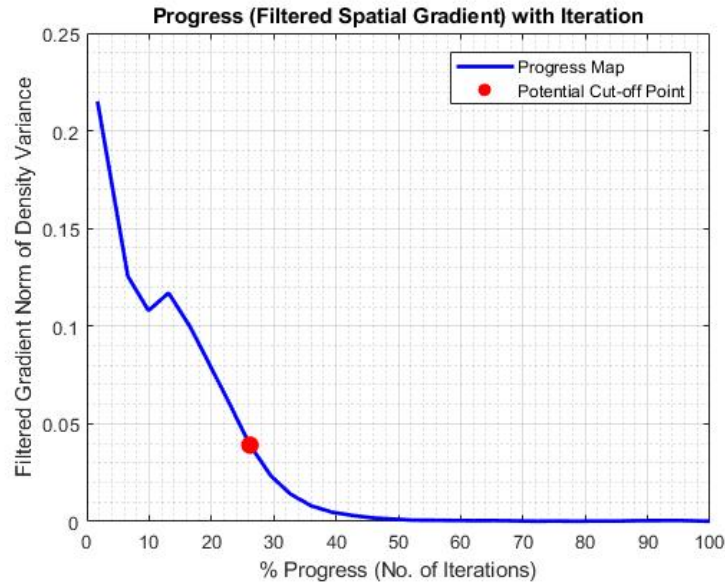

Figure 7: Spatial Gradient Norm Vs Progress (1000 samples)

The mapping captures the trend in the topology optimization process. The spatial gradient tends to zero after first 20-30% of iterations and remains stagnant thereafter. This shows that the structural skeleton is locally resolved within the initial iterations and the change in the actual density values as the algorithm progresses only occurs locally-without altering the global structural outlook too much. This makes the spatial gradient a critical metric, as it can be effectively used to demarcate the point of high structural resolution.

Comparative study of Figure 6 and Figure 7 suggests that the point of nominal change in spatial gradient norm and that of high binary accuracy (above 90%) is almost identical in terms of percentage progress. The only difference between the two is that binary accuracy is based on the final output and requires the optimization process to complete so as to map the data, while gradient data can be incrementally computed making it a more robust and convenient metric to define a possible cutoff point.

Based on this study, during testing for each of the CNN input sample, we progressively mapped the spatial gradient metric individually. The cutoff point for transfer of data to the 3D CNN was defined to be the point where the spatial gradient norm was less than or equal to 0.05 ($|\nabla \tilde{x}| \leq 0.05$). The same strategy can be used as a stopping criterion for the conventional solver during deployment of the network for the given topology optimization problem case.

5. **RESULTS AND DISCUSSION**

The goal of our experiments is to demonstrate how the proposed model and the overall pipeline is useful for solving topology optimization problems efficiently. We compare the performance of our approach with standard linear elasticity solvers in terms of the accuracy of the obtained structure and the average time required for convergence to a solution.

We use two metrics for evaluating the performance of our network:
- Binary Accuracy: Computing equality of binarized values for each voxel in network output with solver output. This is suitable for assessing how the structure obtained by binarizing network output compares against the output from the 'TopOpt' [31] solver.
- Root Mean Squared Accuracy: Calculating square root of average mean squared difference between the float density values for each voxel in the network output and the actual solver output. This is useful for quantifying the network's ability to predict approximately binary density values and how numerically close are those to the actual float values in the output from the 'TopOpt' [31] solver.

The mathematical formulation of the chosen metrics is described as follows [30]:
Let $n_l$ be $l = 0, 1$ be the total number of voxels of class *l*.
Let $\omega_{tp}$ ; $t, p = 0,1$ be the total number of voxels of class *t* predicted to belong to class *p*.

Let $\rho^i_{true}$ and $\rho^i_{pred}$ denote the float density values of discretized element *i* in the ground truth and the network output respectively.
Therefore, the expressions for Binary accuracy and RMS accuracy are given by:

$$Binary\ Accuracy = \frac{\omega_{00} + \omega_{11}}{n_0 + n_1}$$

$$RMS\ Accuracy = 1 - \sqrt{\frac{\sum_{i=1}^{n}(\rho^i_{true} - \rho^i_{pred})^2}{n}}$$

For the test data, we stopped the 'TopOpt' [31] software after specific number of iterations based on the spatial gradient metric. Through comparative studies, we attempt to find the optimal strategy to train the CNN with best accuracy and efficiency. It is expected that a structure which has gone through higher number of iterations in 'TopOpt' [31] solver would give better results for our approach but at the same time, it would also takes more time to get the results. We try to find a balance between number of iterations and the time taken while achieving high accuracy values. Details of the various comparative studies conducted are as follows:

### 5.1 Performance of networks trained on 4 datasets obtained through different input sampling strategies

As previously mentioned, we created four data sets, each of 6000 data samples, using different strategies for density data iteration sampling. The sampling strategies used to create these data sets are tabulated below:

Table 2: Strategies for training CNN

| Training Dataset | Method to train 3D CNN (Density data iteration sampling) |
|---|---|
| 1 | Uniform Distribution |
| 2 | Poisson's Distribution (λ=5) |
| 3 | Poisson's Distribution (λ=10) |
| 4 | Poisson's Distribution (λ=30) |

We trained networks on each of these above datasets and then evaluated the performance of our networks on a common test dataset to compare their performance equitably. The test dataset was obtained by treating the 20th iteration of every problem as input density distribution and also selecting the 15th iteration for computing the density gradient based on the spatial gradient metric. The performance of these networks on the test dataset is depicted in Figure 8 below:

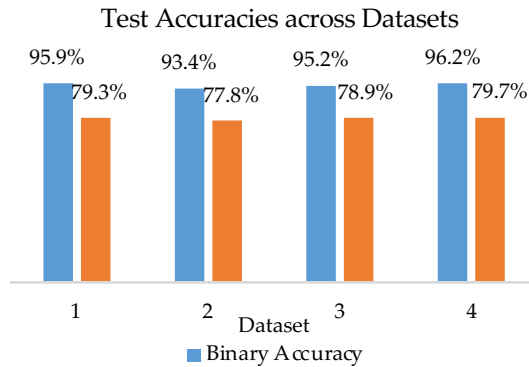

Figure 8: Binary and RMS accuracy comparisons for networks trained on datasets obtained through different sampling strategies

The network trained using sampling strategy of Poisson distribution with λ = 30 yields the best binary accuracy of 96.2% and RMS accuracy of 79.7% and is chosen as the best network. It is notable that the network trained on dataset using uniform sampling is only marginally inferior to the best network. The training inputs for datasets 2 and 3 are poorly resolved structurally, since they are associated with the initial phase of the optimization process. Hence, the network finds it comparatively more difficult to learn the mapping between the input and ground truth structures. This is evident from the marginally lower accuracy values for networks trained using these datasets.

### 5.2 Training on transformed data

We also created a transformed data set, in which we randomly selected 40% of the original training data samples and effected a rotation of their structure along x, y and z axes on a random basis. This data augmentation was a measure for preventing overfitting by introducing random noise in the data. The performance of the network trained on the transformed dataset compared to the network trained on data without transformation was gauged on the same test dataset (without transformations). This yielded the following results:

Table 3: Performance comparison for networks trained using datasets with and without data augmentation

| Performance metric | Network trained on dataset augmented with transformations | | Network trained on dataset without any transformations | |
|---|---|---|---|---|
| | Train | Test | Train | Test |
| RMS Accuracy | **66.7%** | 76.2% | 81.2% | 79.7% |
| Binary Accuracy | 83.4% | 93.3% | 97.7% | **96.2%** |

The insignificant difference between the training and test accuracies for the network trained on data without transformation is indicative that it is not overfitting the data. Furthermore, we infer from the results above that the network is unable to cope with the transformations in the voxel input as evident from the poor training RMS accuracy, though it is still able to perform well on the untransformed test data. Therefore, augmentation of data with transformations is not found to be beneficial in improving performance of the network.

### 5.3 Investigating performance of varying combinations of density data, gradient data and boundary conditions as input

The table below utilizes a Boolean to indicate whether a category of data is part of the input or not and lists the accuracies of the best network (trained on aforementioned dataset 4) using each of these combinations of inputs.

Table 4: Accuracies for different network inputs

| S. No. | Density Data | Gradient Data | Boundary conditions (Forces + constraints) | Binary accuracy | RMS accuracy |
|---|---|---|---|---|---|
| 1 | Yes | No | No | 95.5% | 79.0% |
| 2 | No | Yes | No | 86.4% | 65.8% |
| 3 | No | No | Yes | 73.9% | 58.7% |
| 4 | Yes | Yes | No | **96.2%** | **79.7%** |
| 5 | Yes | No | Yes | 95.3% | 78.9% |
| 6 | No | Yes | Yes | 83.9% | 67.9% |
| 7 | Yes | Yes | Yes | 95.7% | 79.3% |

Several inferences are drawn from the above results:
- The best network is the one which utilizes only the density and gradient data as input to the CNN.
- Inclusion of boundary conditions data along with the density and gradient data as input does not appear to be beneficial for our CNN.
- Absence of the density data is detrimental to the network's performance.
- Using the boundary conditions data alone results in the worst performance. As this data lacks any structural resolution, the network performance is grossly hindered.

*5.4 Investigating performance of varying combinations of iterations for density data & gradient data as input*

The performance of the network improves when provided with higher iteration for density data as input. This is expected because the network receives a higher resolution structure as input which is closer to the ground truth. Furthermore, it is observed that the network's performance is highest when density data is complemented by gradient density data with respect to a prior iteration in close proximity to the density iteration. This provides the network with intuition about the direction of progression of the elemental densities in the structure. However, providing a higher iteration for density data is accompanied by a corresponding increase in computational time since the solver needs to be run for a longer duration. So, the trade-off between spatial accuracy and computational time needs to be balanced well. Keeping in mind the trade-off involved, choosing density data as 20[th] iteration and computing gradient with respect to 15[th] iteration is a compelling strategy since it provides a reasonably good spatial accuracy with low computational time. These findings also directly correlate with the neighborhood spatial gradient cut-off metric defined above. Based on experimentation, choosing this strategy results in nearly 40-45% average reduction in overall computation time.

Table 5: Variation in Binary accuracy upon different iterations for density and gradient data as input to CNN

| Density Iteration \ Gradient Iteration | 5 | 10 | 15 | 20 | 25 | 30 | 35 |
|---|---|---|---|---|---|---|---|
| 5 | - | 81.57% | 80.50% | 80.30% | 80.26% | 80.26% | 80.23% |
| 10 | 91.25% | - | 90.24% | 89.93% | 89.81% | 89.75% | 89.69% |
| 15 | 94.88% | 95.03% | - | 94.46% | 94.24% | 94.12% | 94.05% |
| 20 | 96.08% | 96.17% | **96.22%** | - | 95.90% | 95.77% | 95.68% |
| 25 | 96.65% | 96.73% | 96.80% | 96.80% | - | 96.60% | 96.51% |
| 30 | 96.95% | 97.01% | 97.05% | 97.08% | 97.06% | - | 96.97% |
| 35 | 97.10% | 97.14% | 97.18% | 97.20% | 97.20% | 97.19% | - |

Table 6: Change in RMS accuracy depending on different iterations for density and gradient data as input to CNN

| Density Iteration \ Gradient Iteration | 5 | 10 | 15 | 20 | 25 | 30 | 35 |
|---|---|---|---|---|---|---|---|
| 5 | - | 69.04% | 68.00% | 67.80% | 67.73% | 67.70% | 67.67% |
| 10 | 75.39% | - | 74.45% | 74.05% | 73.89% | 73.81% | 73.76% |
| 15 | 78.28% | 78.43% | - | 77.80% | 77.54% | 77.40% | 77.32% |
| 20 | 79.57% | 79.66% | **79.71%** | - | 79.32% | 79.14% | 79.03% |
| 25 | 80.19% | 80.25% | 80.32% | 80.31% | - | 80.08% | 79.96% |
| 30 | 80.52% | 80.57% | 80.60% | 80.63% | 80.62% | - | 80.50% |
| 35 | 80.70% | 80.72% | 80.74% | 80.76% | 80.79% | 80.79% | - |

6. **CONCLUSIONS**

The findings of the paper show that a deep learning framework can indeed be deployed to predict final outputs of topology optimization from intermediate structural inputs. We found the technique to be reliable even for 3D topology optimization problems. In fact, the reasonably good RMS accuracies for the network output demonstrate the ability of the 3D CNN to learn the mapping from initial to the final density distributions. Moreover, the identical nature of final outputs (after thresholding) of the CNN and the FEA solver, which is evident from the high binary accuracy values, validates the practical applicability of the proposed methodology through the use of machine learning techniques.

We conclude that the network with the best performance on the test set is the one which is trained on the dataset with density data iteration sampled using the Poisson distribution with $\lambda = 30$. The optimal input composition to the CNN is a combination of density distribution and density gradient without data for forces and boundary conditions. The model does not overfit the training data, as inferred by the marginal difference between the binary and RMS accuracies on the training and the test data. So, the input is not augmented with random transformations.

The optimum computational strategy of using spatial gradient of filtered densities (section 4.4.2) deduced in our research shows that interpolating the final output using our 3D CNN from the initial iterations (section 5.4) obtained from the 'TopOpt' [31] solver, offers a 40% reduction in time over the conventional approach of using the solver alone. This strategy results in the best balance of spatial accuracy and computational time and yields an average binary accuracy of 96.2% and average RMS accuracy of 79.7% on the test set.

There exist numerous possibilities for future work in this domain. Some critical ones are listed below:
- The possibility of a better way to incorporate forces and constraints that help the network learn better.
- Enabling the user to specify the volume fraction providing more control over the problem formulation.
- A better penalization function to drive the density values towards convergence.
- Training the network on more complex problems apart from the MBB beam.
- Feeding the sub-optimal output from the network back to the optimization software to continue convergence towards a binary solution.
- Making sure that there are no islands created in the structure predicted by the network.

Visual comparison of the network output (prediction) against the conventional solver output (ground truth) for some sample structures is depicted in Table 7 below. Refer to Appendix C, for a consolidated list of network output comparisons which validates its performance.

Table 7: Comparison of 'TopOpt' [31] output and CNN output at 20$^{th}$ input iteration

| STRUCTURE | BOUNDARY CONDITIONS | ACCURACY | INPUT | CNN PREDICTION | GROUND TRUTH |
|---|---|---|---|---|---|
| 1 | 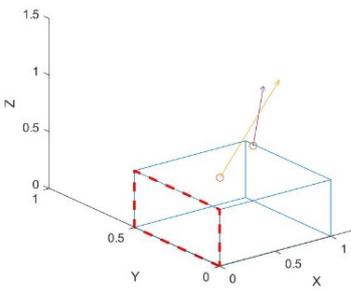 | WITHOUT THRESHOLDING | | | |
| | | RMS = 84.35 | 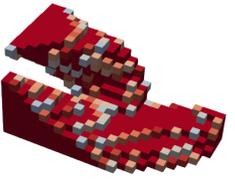 | 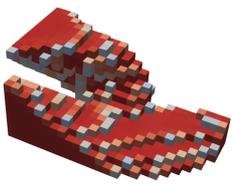 | 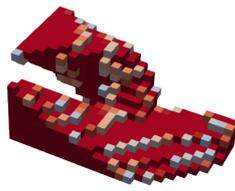 |
| | | WITH THRESHOLDING AT 0.5 | | | |
| | | BINARY = 98.84 | 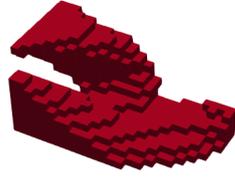 | 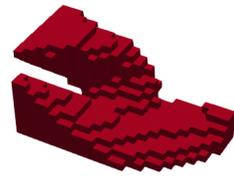 | 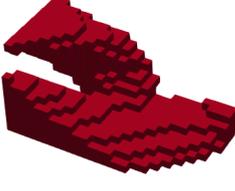 |
| 2 | 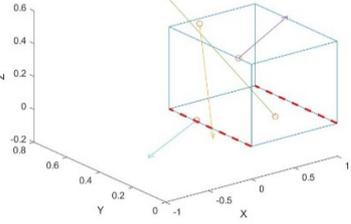 | WITHOUT THRESHOLDING | | | |
| | | RMS = 84.32 | 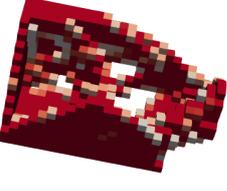 | 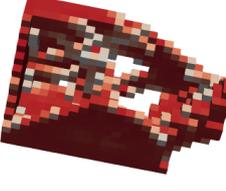 | 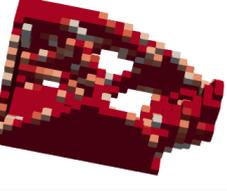 |
| | | WITH THRESHOLDING AT 0.5 | | | |
| | | BINARY = 97.36 | 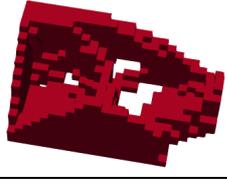 | 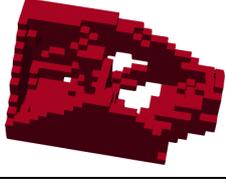 | 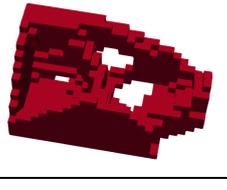 |

**ACKNOWLEDGEMENT**

We would like to express our gratitude to Dr. Erva Ulu for his guidance and support on our research work, which proved to be a landmark effort towards the success of our work. We would also like to extend our thanks to all those who have directly and indirectly contributed to this research by providing valuable suggestions and encouragement which gave us inspiration to lead the work to completion.

**NOTE:** THIS RESEARCH DID NOT RECEIVE ANY SPECIFIC GRANT FROM FUNDING AGENCIES IN THE PUBLIC, COMMERCIAL, OR NOT-FOR-PROFIT SECTORS

## APPENDIX

### A. Training progress

We use the best network to train on dataset comprising of 4500 data points sampled using strategy of Poisson distribution with $\lambda = 30$ for 30 epochs. The averaged accuracy across epochs plot in figure A.1 depicts the progress of the network. It can be observed that the accuracies are monotonically increasing. The accuracy values are indicative of the network's ability to identify the features of the data and predict the final structure to a high extent.

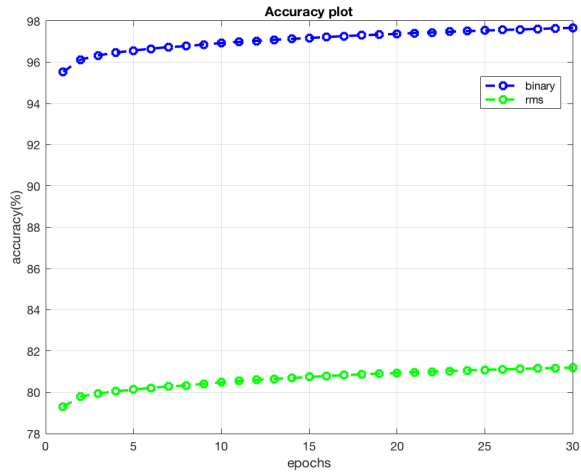

Figure A.1: Averaged accuracy (binary and RMS) over each epoch

Figure A.2 represents the averaged total loss over each epoch. The loss monotonically decreases through the epochs albeit decreasing at a slow pace.

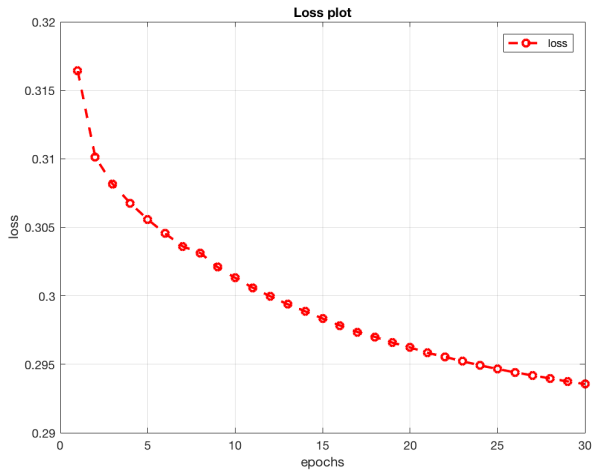

Figure A.2: Averaged total loss over each epoch

While the high accuracy after the first epoch can be surprising, it represents the accuracy after 4500 updates of the weights of the network since we have used Stochastic Gradient Descent (SGD) with momentum as our optimizer without mini-batch. A mini-batch of size 10 for training resulted in an inferior performance and much slower training speeds.

Moreover, the process mapping indicates that the structure achieves a high binary accuracy in the initial stages (~15-20% iterations) of the optimization process. The training accuracy and loss progress plots for the first epoch are depicted in Figure A.3 and Figure A.4 respectively.

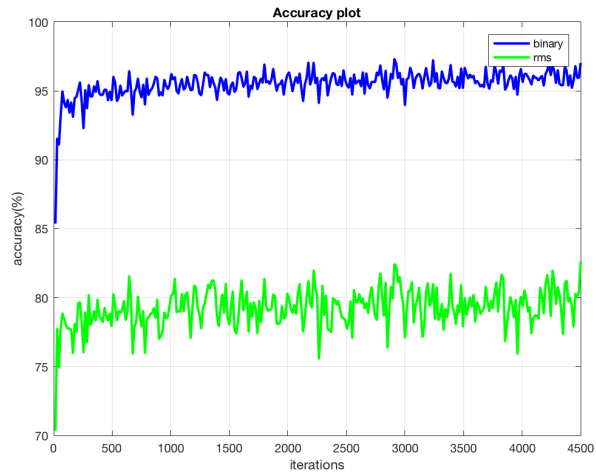

Figure A.3: Averaged accuracy (binary and RMS) over the first epoch

It can be observed that the plots appear noisy and oscillatory, which is typical for SGD optimizer. However, the averaged loss over the entire epoch decreases and the averaged accuracies over the epoch increase.

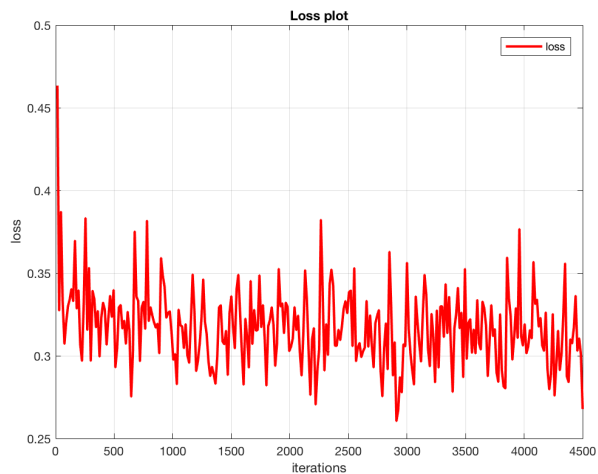

Figure A.4: Averaged loss over the first epoch

## B. Process Mapping

**Gradient Norm:**

The iteration-wise gradient of the material density data was computed for all the 1000 test samples. Mapping of the Frobenius norm of these 3D gradient arrays against percentage progress in terms of iterations is shown in figure B.1.

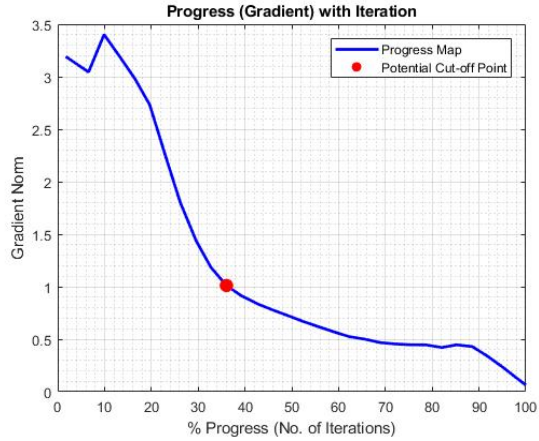

Figure B.1: Gradient Norm Vs Progress (1000 samples)

The gradient norm is observed to continually decrease with a steep slope for initial few iterations followed by a state of gradual change tending to zero until convergence. This trend is in line with the standard gradient-based numerical algorithms like Optimality Criteria (OC) or Sequential Quadratic Programming (SQP) among others typically used for topology optimization. As the gradient-norm keeps on monotonously decreasing, it does not capture the essence of high spatial resolution observed early in the process since the material density values keep changing – although locally and marginally in the latter stages- and hence the gradient norm never becomes zero until convergence.

## C. Network Outputs

Table C.1: An exhaustive set of examples as predicted by our network and ground truth obtained via solver with network accuracy of prediction.

| Network Prediction (Float Densities) | Ground Truth (Float Densities) | Network Prediction (After Thresholding) | Ground Truth (After Thresholding) | Accuracy |
|---|---|---|---|---|
| | | | | RMS = 83.83% <br> Binary = 98.23 % |
| | | | | RMS = 83.95 % <br> Binary = 96.15 % |
| 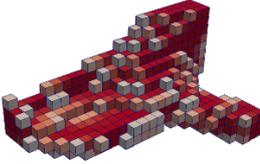 | 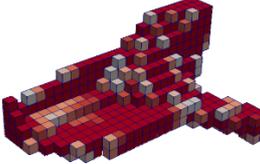 | 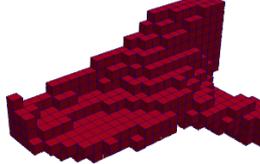 | 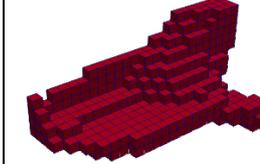 | RMS = 72.90 %<br>Binary = 95.83 % |
| 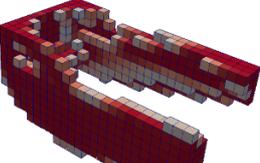 | 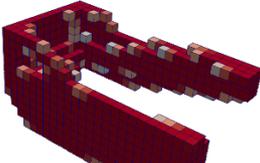 | 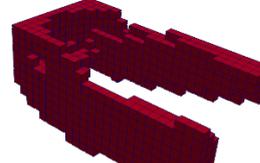 | 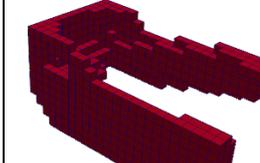 | RMS = 73.91 %<br>Binary = 94.70 % |
| 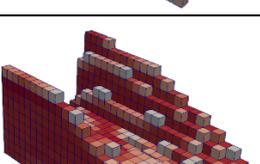 | 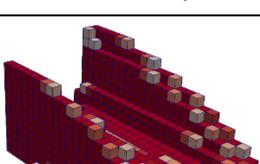 | 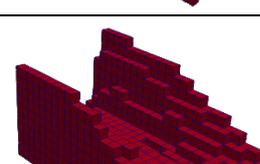 | 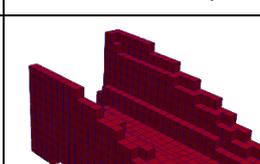 | RMS = 78.99 %<br>Binary = 93.14 % |
| 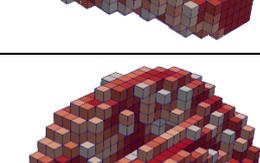 | 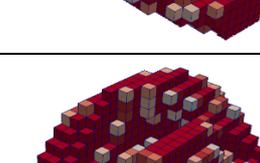 | 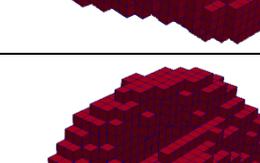 | 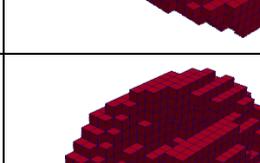 | RMS = 78.42 %<br>Binary = 93.17 % |
| 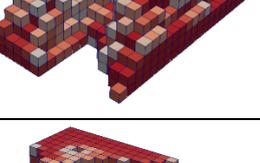 | 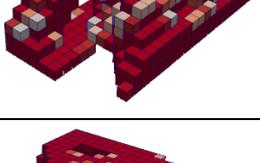 | 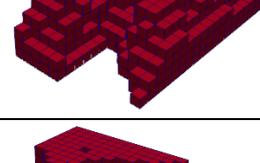 | 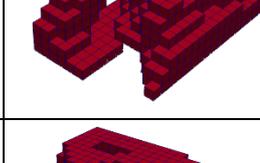 | RMS = 81.80 %<br>Binary = 93.92 % |
| 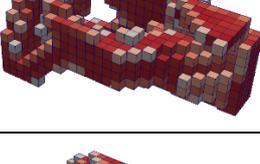 | 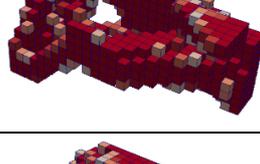 | 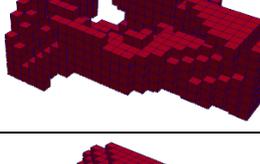 | 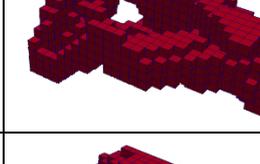 | RMS = 85.17 %<br>Binary = 99.01 % |
| 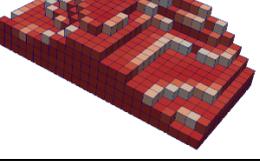 | 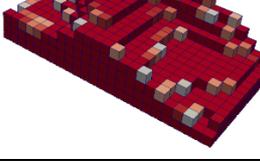 | 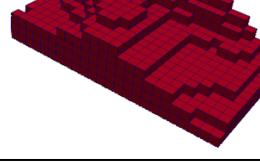 | 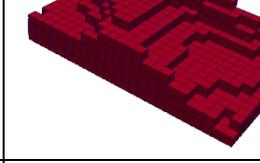 | RMS = 83.02 %<br>Binary = 97.59 % |
| 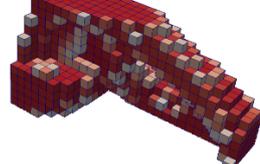 | 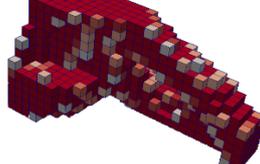 | 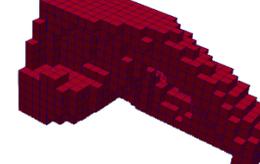 | 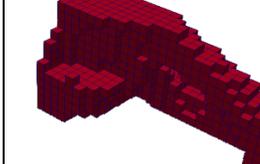 | RMS = 80.11 %<br>Binary = 98.81 % |
| 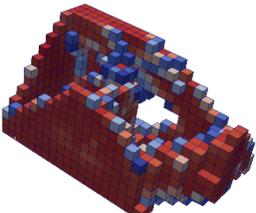 | 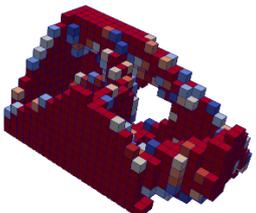 | 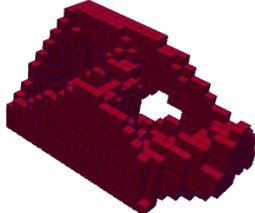 | 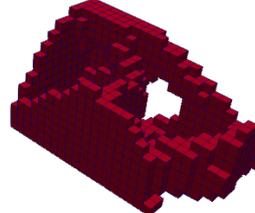 | RMS = 83.69 %<br>Binary = 96.58 % |
| 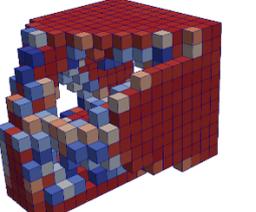 | 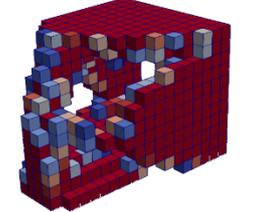 | 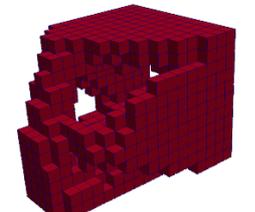 | 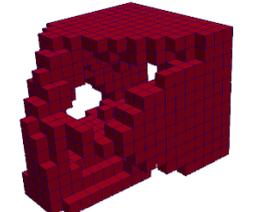 | RMS = 80.02 %<br>Binary = 98.00 % |
| 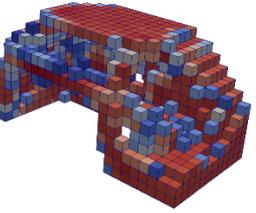 | 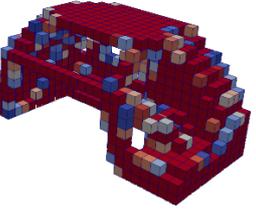 | 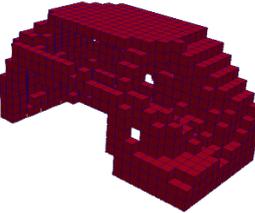 | 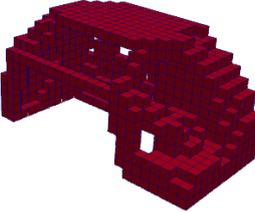 | RMS = 84.13 %<br>Binary = 96.23 % |
| 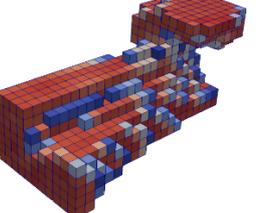 | 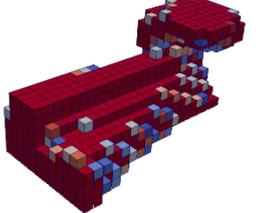 | 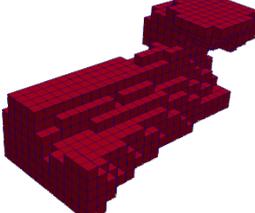 | 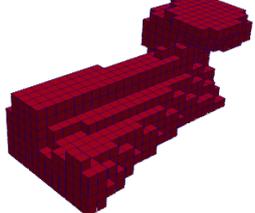 | RMS = 85.58 %<br>Binary = 97.59 % |
| 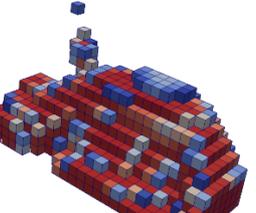 | 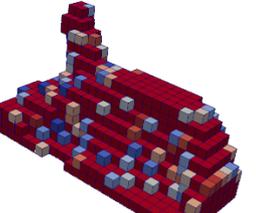 | 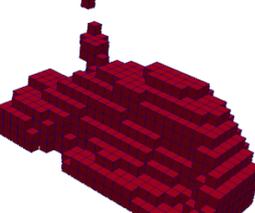 | 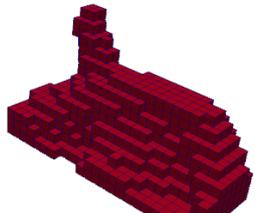 | RMS = 77.79 %<br>Binary = 95.57 % |